\documentclass[a4paper, 10 pt, conference]{hsmr} 
\usepackage[utf8]{inputenc}
\IEEEoverridecommandlockouts                       
\usepackage{mathtools}
\usepackage{geometry}
\usepackage{newtxmath,newtxtext}
\usepackage{authblk}
\usepackage{pgfplots}
\usepackage{graphicx}
\usepackage{bm}
\usepackage{cite}

\usepackage{newtxtext,newtxmath}

\pgfplotsset{compat=newest} 
 
\title{\LARGE \bf
Toward a Millimeter-Scale Tendon-Driven Continuum Wrist with Integrated Gripper for Microsurgical Applications} 

\author{\LARGE Alexandra Leavitt$^{1,2}$}
\author{\LARGE Ryan Lam$^{1,2}$}
\author{\LARGE Nichols Crawford Taylor$^{1,2}$}
\author{\LARGE \\Daniel S. Drew$^{1,3}$}
\author{\LARGE Alan Kuntz$^{1,2}$\thanks{This work was supported in part by the National Science Foundation under Award \# 2133027. A. Leavitt and R. Lam were also supported in part by funding from the Undergraduate Research Opportunities Program at the University of Utah. This work was performed in part at the Utah Nanofab. The authors appreciate the support of the Nanofab staff and facility. We also thank the groups of Dr. D. Caleb Rucker and Dr. Jake
J. Abbott for valuable discussions.}} 

\affil{\Large\textit{$^{1}$Robotics Center, $^{2}$Kahlert School of Computing, }\\ \Large\textit{$^{3}$Department of Electrical and Computer Engineering, University of Utah}\\
\Large\textit{alan.kuntz@utah.edu}}

\begin{document}

\maketitle
\thispagestyle{empty}
\pagestyle{empty}

\section*{INTRODUCTION}
Microsurgery, wherein surgeons operate on extremely small structures frequently visualized under a microscope, is a particularly impactful yet challenging form of surgery.
Robot assisted microsurgery has the potential to improve surgical dexterity and enable precise operation on such small scales in ways not previously possible~\cite{Mattos2016_SMW,Zhang2022_MIR}.
Clinical applications of microsurgery include intraocular surgery, fetal surgery, otology, laryngeal surgery, neurosurgery, and urology.

Intraocular microsurgery is a particularly challenging domain~\cite{Ahronovich2021_AdvT,Iordachita2022_PIEEE}.
Challenges arise, in part, due to the lack of dexterity that is achievable with rigid instruments inserted through the eye.
The insertion point introduces a remote center of motion constraint that prevents control over a tool-tip's full position and orientation for conventional, straight instruments.
Continuum robots based on concentric tubes~\cite{Lin2015_IEMBC}, magnetic actuation~\cite{Charreyron2021_TBME}, and tendon-actuated stacked disks~\cite{Jinno2021_JMRR,Jinno2022_ICRA} have been proposed for intraocular microsurgery to overcome this constraint, but are frequently limited in their local curvatures---an important consideration in constrained spaces.

\begin{figure}
\centering
\includegraphics[width=\columnwidth]{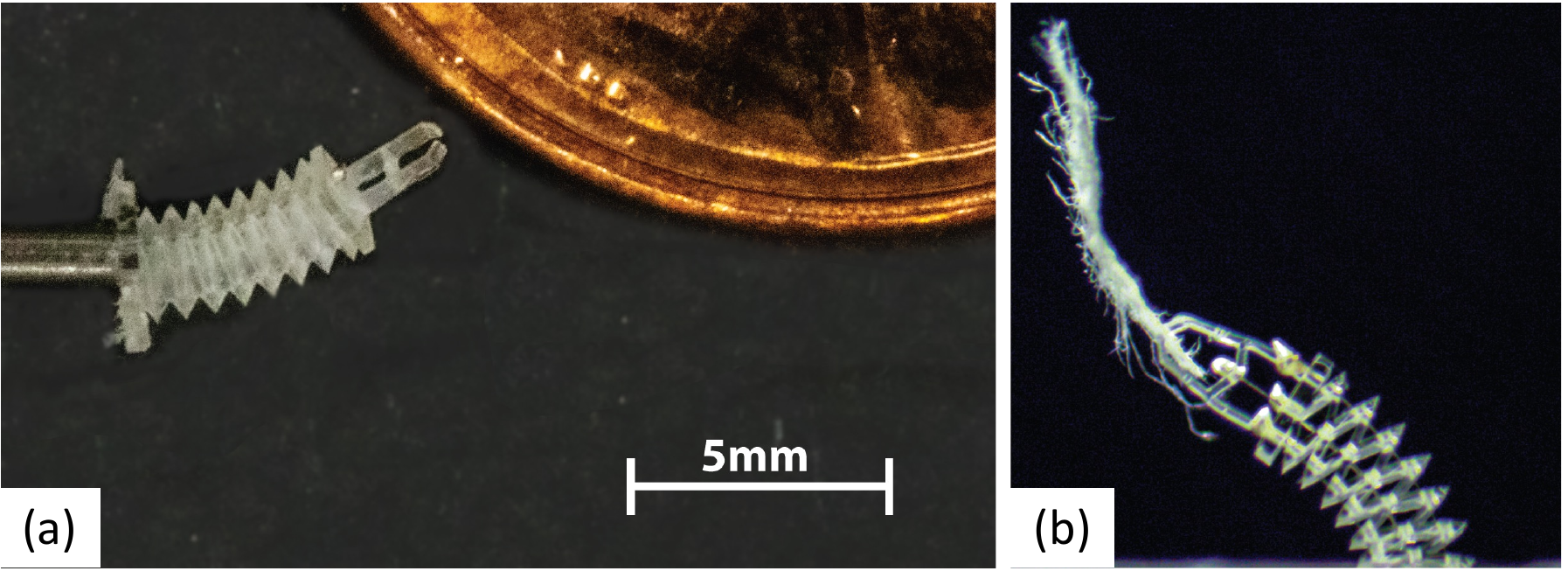}
\vspace*{-5mm}
\caption{(a) The wrist/gripper presented in this work mounted on a stainless steel capillary tube shown next to the edge of a US 1 cent coin for scale.
(b) The gripper enables the manipulation of loads while bending. Here, microfiber paper is used as a proxy for tissue.}
\label{fig:overview}
\vspace{-2mm}
\end{figure}

Inspired by these works, and to take steps toward overcoming limited local curvature, we present a new design for a millimeter-scale, dexterous tendon-driven continuum wrist and gripper, intended for microsurgery applications.
The device is created via a state-of-the-art two-photon-polymerization (2PP) microfabrication technique.
The 2PP 3D printing method enables construction via a flexible material, with complex internal geometries and critical features at the micron-scale (Fig.~\ref{fig:overview}).

The wrist is composed of a stacked rhombus shape, first proposed at macro scale by Childs et. al.~\cite{Childs2021_FRAI}.
This design introduces torsional rigidity as a byproduct of its geometry.
We leverage this design not for its torsional rigidity (although we envision that will aid modeling in future work), but rather due to the fact that the extruded nature of its geometry lends itself to 2PP sub-millimeter scale 3D printing.
Building on this concept, we miniaturise the design and integrate a flexible gripper.
The wrist has three tendons routed down its length, spaced approximately evenly around its circumference which, when actuated by small-scale linear actuators, enable bending in any plane.
The gripper is actuated by a fourth tendon routed down the center of the wrist.

We characterize the wrist's bend-angle, achieving $>$90\textdegree{} bend in both axes.
We demonstrate out-of-plane bending as well as the ability to grip while actuated.

Our integrated tendon-driven continuum wrist/gripper design and meso-scale assembly techniques have the potential to enable small-scale robots with more dexterity than has been previously demonstrated.
Such a wrist could improve surgeon capabilities during teleoperation with the potential to improve patient outcomes in a variety of surgical applications, including intraocular surgery.

 \begin{figure}[t]
    \centering
    \includegraphics[width=\columnwidth]{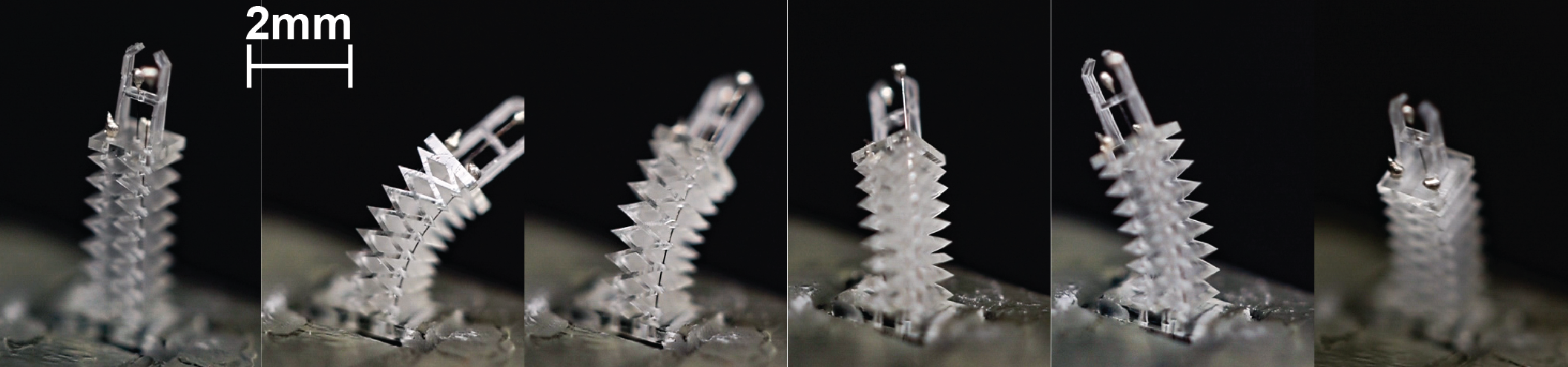}
  \vspace*{-5mm}
  \caption{Sequential images of the wrist tracing a circular trajectory via coordinated actuation of the three  tendons.}
  \label{fig:large-circle}
  \vspace{-6mm}
\end{figure}

\section*{MATERIALS AND METHODS}
The wrist features a square cross section with side length of 1.25\,mm and total length of 3.75\,mm.
The wrist comprises repeat hollow rhombohedral unit cells (see Fig.~\ref{fig:overview}), as in Childs et. al.~\cite{Childs2021_FRAI}. Flexural bending about each axis results from elastic deformations of the unit cells, which can be individually modeled as a parallel set of conjoined thin plates.
The unit cells are torsionally stiff because of the relatively small thickness of the thin plate sections relative to their width and length.

As a proof-of-concept, we designed a simple gripper which can be controlled by pulling a centrally-routed tendon (Fig.~\ref{fig:overview}b). The tendon symmetrically bends the simply supported ends of the center plate, bringing the gripper jaws closer until they make contact.

The wrist and integrated gripper, with a total length of 5.47\,mm, are fabricated using a 2PP printing process (Nanoscribe 3D Photonic Pro GT), enabling direct printing of three-dimensional structures with sub-micrometer resolution, using the semi-rigid IP-Q photoresin ($E\approx$5\,GPa, $\nu\approx$0.35).

Tendons (25\,$\mu$m diameter tungsten) are manually threaded through the channels in the wrist.
Two-part silver epoxy adhesive (MG Chemicals 8331D) is used for capping the tendons.
Tendons are attached to linear actuators with 20\,mm stroke, 18\,N rated force, and $\approx$ 100\,$\mu$m minimum step size (Actuonix PQ12-R).

\vspace{-0.5mm}
\section*{RESULTS}
\vspace{-0.5mm}

We qualitatively demonstrate out-of-plane bending of the wrist and use of the gripper.
Actuating each of the three bending tendons in a coordinated fashion causes the wrist to bend in a 3D circular arc (Fig.~\ref{fig:large-circle}). In Fig.~\ref{fig:overview}b we show grasping of a piece of tissue paper having externally actuated both the gripper tendon and a bending tendon.

We next quantitatively characterize the wrist's ability to bend in each of its orthogonal axes as a function of tendon displacement.
In this experiment, we printed the wrist structure (without the gripper) with four bending tendons, located on the wrist such that there were two tendons antagonistically placed in both of the wrist's primary axes.
We placed the wrist in front of a grid printed with 1\,mm~x~1\,mm cells to optically measure bend angle through a stereo microscope.

\begin{figure}[t]
    \centering
    \includegraphics[width=\columnwidth]{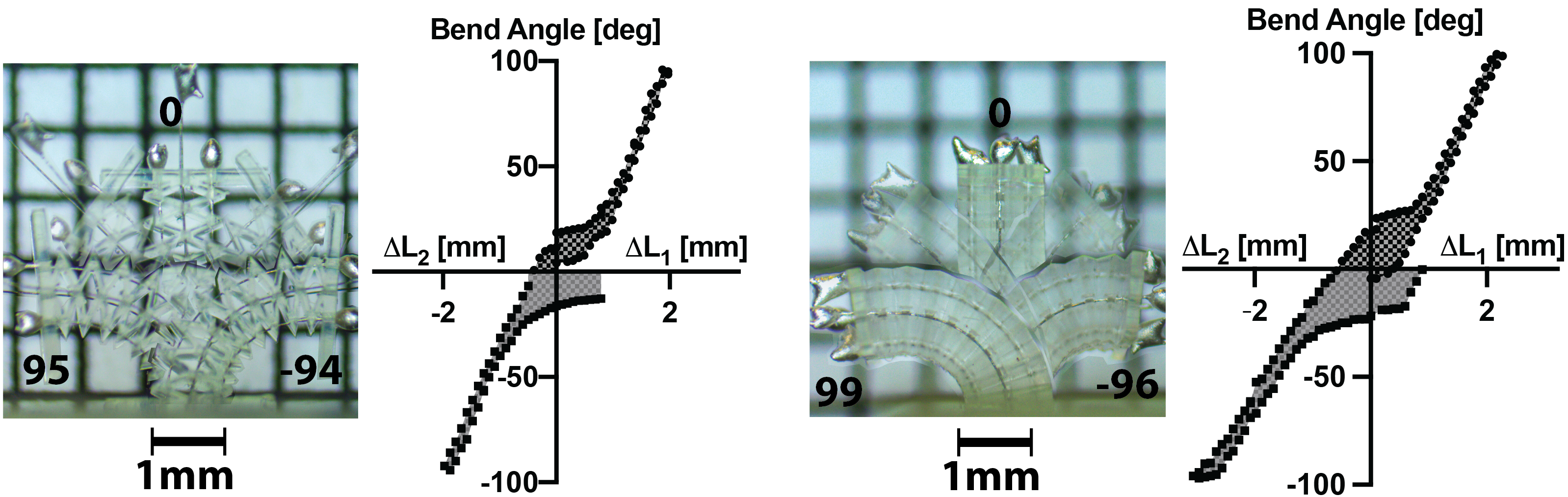}
     \vspace*{-7mm}
  \caption{Bend angles with hysteresis curves as a function of tendon displacement ($\Delta \textrm{L}_1$ and $\Delta \textrm{L}_2$) of the in-plane tendons in both primary axes. The wrist exhibits $\geq$ 90\textdegree{} bending in both primary axes. (Left) the wrist as viewed from the front. (Right) the wrist as viewed from the side. Multiple bend angles are superimposed in each figure.}
  \label{fig:large-bends}
  \vspace{-5mm}
\end{figure}

For each of the two axes we actuated one of the bending tendons, measuring tendon displacement (via actuator encoders) and bend angle until the angle was 90\textdegree{} or more.
Note that this was an artificially imposed limit and the wrist could be actuated further.
We then released the tendon tension, measuring forward displacement of the actuator until we had returned to the initial displacement length.
Next we actuated the appropriate antagonistic tendon to bend the wrist in the opposite direction in plane repeating the measurements as described above.
In Fig.~\ref{fig:large-bends} we show overlaid images of the wrist bending in both directions for both axes.
We include a hysteresis curve for each bending axis quantifying bend angle versus the displacement of the two tendons in the bending plane ($\Delta \textrm{L}_1$ and $\Delta \textrm{L}_2$).
We achieve $>$90\textdegree{} bending in all four directions.

\vspace{-0.5mm}
\section*{DISCUSSION}
\vspace{-0.5mm}
Our 2PP fabricated device demonstrates large bend angles ($>$90\textdegree{}) in a small form factor (3.75\,mm wrist length).
We demonstrate high curvature in the wrist with values between 2.1 and 2.3\,mm radius of curvature.
However, static friction in the tendon channels introduces significant hysteresis.
We envision the wrist as part of a robot being teleoperated with closed-loop feedback mapping human input to bend angle; ramifications of the hysteresis in that setting will need to be studied, particularly with antagonistic tendons potentially closing the hysteresis curve.
Statistical evaluation of tension/force and achievable bending, and bending beyond our artificially-limited $>$90\textdegree{} angle, remain for future work.
Further, we do not yet have a model mapping tendon tension or displacement to bend angle, which is needed for closed-loop control.

This work represents our first demonstration of 2PP fabricated wrists at these scales. In future work we intend to both address the above limitations and to further reduce the scale of the device.
We envision a large number of potential clinical uses for such a device.
Our primary intended use case is to integrate our wrist with a teleoperated intraocular microsurgery robot.
\nocite{*}
\bibliographystyle{IEEEtran}
\bibliography{main}

\end{document}